\documentclass[10pt,twocolumn,letterpaper]{article}

\usepackage{cvpr}
\usepackage{times}
\usepackage{graphicx}
\usepackage{rotating}
\usepackage{multirow}
\usepackage{epstopdf}
\usepackage{bibspacing}
\usepackage{bbm}

\usepackage{subfig}

\newcommand{\secref}{Sec~\ref}{}
\newcommand{\eqnref}{Eq~\eqref}{}
\newcommand{\tabref}{Table~\ref}{}
\newcommand{\figref}{Fig~\ref}{}

\newcommand{\changes}{}
\newcommand{\cc}{}
\newcommand{\KG}{}
\usepackage{amssymb}

\usepackage{amsmath,amsfonts}
\usepackage{amsthm,amsopn}

\usepackage{bm} %

\newcommand{\ProbOpr}[1]{\mathbb{#1}}
\newcommand{\expect}[2]{%
\ifthenelse{\equal{#2}{}}{\ProbOpr{E}_{#1}}
{\ifthenelse{\equal{#1}{}}{\ProbOpr{E}\left[#2\right]}{\ProbOpr{E}_{#1}\left[#2\right]}}} %
\DeclareMathOperator*{\argmax}{arg\,max}
\DeclareMathOperator*{\argmin}{arg\,min}

\usepackage{booktabs}
\newcommand{\ra}[1]{\renewcommand{\arraystretch}{#1}}

\usepackage[font=small,labelfont=bf]{caption}

\cvprfinalcopy %

\def\cvprPaperID{582} %

\ifcvprfinal\fi

\title{{Slow and steady feature analysis:\\ higher order temporal coherence in video}}

\author{Dinesh Jayaraman\\
UT Austin\\
{\tt\small dineshj@cs.utexas.edu}
\and
Kristen Grauman\\
UT Austin\\
{\tt\small grauman@cs.utexas.edu}
}

\begin{document}

\maketitle

\begin{abstract}
  \vspace{-0.1in}

How can unlabeled video augment visual learning? Existing methods perform ``slow" feature analysis, encouraging the representations of temporally close frames to exhibit only small differences.  While this standard approach captures the fact that high-level visual signals change slowly over time, it fails to capture \emph{how} the visual content changes.  We propose to generalize slow feature analysis to ``steady" feature analysis.  The key idea is to impose a prior that higher order derivatives in the learned feature space must be small.  To this end, we train a convolutional neural network with a regularizer on tuples of sequential frames from unlabeled video. It encourages feature changes over time to be smooth, i.e., similar to the most recent changes.  Using five diverse datasets, including unlabeled YouTube and KITTI videos, we demonstrate our method's impact on object, scene, and action recognition tasks.  We further show that our features learned from unlabeled video can even surpass a standard heavily supervised pretraining approach.
\end{abstract}

\vspace{-0.2in}
\section{Introduction}\label{sec:intro}

Visual feature learning with deep neural networks has yielded dramatic gains for image recognition tasks in recent  years~\cite{krizhevsky-nips2012,simonyan2014}.  While the main techniques involved in these methods have been known for some time, a key factor in their recent success is the availability of large human-labeled image datasets like ImageNet~\cite{deng2009imagenet}. Deep convolutional neural networks (CNNs) designed for image recognition typically have millions of parameters, necessitating notoriously large training databases to avoid overfitting. %

Intuitively, however, visual learning should not be restricted to sets of category-labeled exemplars.  Taking human learning as an obvious example, children build up visual representations through constant observation and action in the world.  This hints that machine-learned representations would also be well served to exploit long-term \emph{video} observations, even in the absence of deliberate labels.  Indeed, researchers in cognitive science find that \emph{temporal coherence} plays an important role in visual learning.  For example, altering the natural temporal contiguity of visual stimuli hinders translation invariance in the inferior temporal cortex~\cite{dicarlo-science2008}, and functions learned to preserve temporal coherence share behaviors observed in complex cells of the primary visual cortex~\cite{berkes-2005}.

Our goal is to exploit unlabeled video, as might be obtained freely from the web, to improve visual feature learning.  In particular, we are interested in improving learned image representations for visual recognition tasks.

\begin{figure}[t]
  \centering
    \includegraphics[width=1\linewidth]{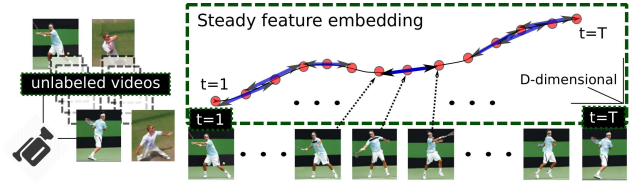}
    \caption{From unlabeled videos, we learn ``steady features'' that exhibit consistent feature transitions among sequential frames.}
  \label{fig:concept}
  \vspace{-0.15in}
\end{figure}

Prior work leveraging video for feature learning focuses on the concept of \emph{slow feature analysis} (SFA).  First formally proposed in~\cite{Wiskott2002}, SFA exploits temporal coherence in video as ``free" supervision to learn image representations invariant to small transformations.  In particular, SFA encourages the following property: in a learned feature space, temporally nearby frames should lie close to each other, \ie for a learned representation $\mathbf{z}$ and adjacent video frames $\bm{a}$ and $\bm{b}$, one would like $\mathbf{z}(\bm{a}) \approx \mathbf{z}(\bm{b})$.  The rationale behind SFA rests on a simple observation: high-level semantic visual concepts associated with video frames typically change only gradually as a function of the pixels that compose the frames. Thus, representations useful for recognizing high-level concepts are also likely to possess this property of ``slowness''.  Another way to think about this is that scene changes between temporally nearby frames are usually small and represent label-preserving transformations.  A slow representation will tolerate minor geometric or lighting changes, which is essential for high-level visual recognition tasks.  The value of exploiting temporal coherence for recognition has been repeatedly verified in ongoing research, including via modern deep convolutional neural network implementations~\cite{Mobahi2009,bergstra-nips2009,drlim,zou2012deep,Goroshin2014,wang-gupta-2015}.

However, existing approaches require only that high-level visual signals change slowly over time.  Crucially, they fail to capture \emph{how} the visual content changes over time. In contrast, our idea is to incorporate the \emph{steady visual dynamics} of the world, learned from video. For instance, if trained on videos of walking people, slow feature-based approaches would only require that images of people in nearby poses be mapped \emph{close} to one another. In contrast, we aim to learn a feature space in which frames from a novel video of a walking person would follow a smooth, predictable trajectory.  {A learned \emph{steady} representation capturing such dynamics would be influenced not only by object motions,} but also other types of visual transformations.  For instance, {it would capture how} colors of objects in the sunlight change over the course of a day, or how the views of a static {scene} change as a camera moves around it.

To this end, we propose {\emph{steady feature analysis}}---a generalization of slow feature learning.  The key idea is to impose higher order temporal constraints on the learned visual representation.  Beyond encouraging temporal coherence \ie, \emph{small feature differences} between nearby frame pairs, we would like to encourage \emph{{consistent} feature transitions} {across} sequential frames. In particular, to preserve second order slowness, we look at triplets of temporally close frames $\bm{a}$, $\bm{b}$, $\bm{c}$, and encourage the learned representation to have $\mathbf{z}(\bm{b})-\mathbf{z}(\bm{a}) \approx \mathbf{z}(\bm{c})-\mathbf{z}(\bm{b})$.
We develop a regularizer that uses contrastive loss over tuples of frames to achieve such mappings with CNNs.
Whereas {slow feature learning} insists that the features not change too quickly, the proposed steady learning insists that---in whichever way the features are evolving---they \emph{continue to evolve in that same way} in the immediate future.  See Figure~\ref{fig:concept}.

We hypothesize that higher-order temporal coherence could provide a valuable prior for recognition by embedding knowledge of the rich dynamics of the visual world into the feature space.
We empirically verify this hypothesis using five datasets for a variety of recognition tasks, including object instance recognition, large-scale scene recognition, and action recognition from still images.  In each case, by augmenting a small set of labeled exemplars with unlabeled video, the proposed method generalizes better than both a standard discriminative CNN as well as a CNN regularized with existing slow temporal coherence metrics~\cite{drlim,Mobahi2009}.  %
Our results reinforce that unsupervised feature learning from unconstrained video is an exciting direction, with promise to offset the large labeled data requirements of current state-of-the-art computer vision approaches by exploiting virtually unlimited unlabeled video.

\vspace{-0.05in}
\section{Related Work}\label{sec:related}
\vspace{-0.05in}

To build a robust object recognition system, the image representation must incorporate some degree of \emph{invariance} to changes in pose, illumination, and appearance.  While invariance can be manually crafted, such as with spatial pooling operations or gradient descriptors, it may also be learned.  One approach often taken in the convolutional neural network (CNN) literature is to pad the training data by systematically perturbing raw images with label-preserving transformations (e.g., translation, scaling, intensity scaling, etc.)~\cite{simard2003best,vincent2008extracting,Dosovitskiy2014}.  A good representation will ensure that the jittered versions originating from the same content all map close by in the learned feature space.

In a similar spirit, unlabeled video is an appealing resource for recovering invariance.  The simple fact that things typically cannot change too quickly from frame to frame makes it possible to harvest sets of sequential images whose learned representations ought not to differ substantially.  Slow feature analysis (SFA)~\cite{Wiskott2002,hurri} leverages this notion to learn features from temporally adjacent video frames.

Recent work uses CNNs to explore the power of learning {slow features, also referred to as} ``temporally coherent'' features~\cite{Mobahi2009,bergstra-nips2009,zou2012deep,Goroshin2014,wang-gupta-2015}.  The existing methods either produce a holistic image embedding~\cite{Mobahi2009,bergstra-nips2009,Goroshin2014,drlim}, or else track local patches to learn a localized representation~\cite{zou2012deep,willetal,wang-gupta-2015}.  Most methods exploit the learned features for object recognition~\cite{Mobahi2009,zou2012deep,bergstra-nips2009,wang-gupta-2015}, while others employ them for dimensionality reduction~\cite{drlim} or video frame retrieval~\cite{Goroshin2014}.  In~\cite{Mobahi2009}, a standard deep CNN architecture is augmented with a temporal coherence regularizer, then trained using video of objects on clean backgrounds rotating on a turntable.  The method of~\cite{bergstra-nips2009} builds on this concept, proposing the use of decorrelation to avoid trivial solutions to the slow feature criterion, with applications to handwritten digit classification.  The authors of~\cite{Goroshin2014} propose injecting an auto-encoder loss and explore training with unlabeled YouTube video. Building on SFA subspace ideas~\cite{Wiskott2002}, researchers have also examined slow features for action recognition~\cite{zhang-pami2012}, facial expression analysis~\cite{zafeiriou-iccv2013}, \changes{\cc{future prediction~\cite{Vondrick2015},}} and temporal segmentation~\cite{nater-bmvc2011,liwicki-accv2012}.

Related to all the above methods, we aim to learn features from unlabeled video.  However, whereas all the past work aims to preserve feature \emph{slowness}, our idea is to preserve higher order feature \emph{steadiness}.  Our learning objective is the first to move beyond adjacent frame neighborhoods, requiring not only that sequential features change gradually, but also that they change in a similar manner in adjacent time intervals.

Another class of methods learns \emph{transformations}~\cite{Michalski2014, Memisevic2013,Ranzato2014}.  Whereas the above feature learning methods (and ours) train with unlabeled video spanning various unspecified transformations, these methods instead train with pairs of images for which the transformation is known and/or consistent.  Then, given a novel input, the model can be used to predict its transformed output.  %
Rather than use learned transformations for extrapolation like these approaches, our goal is to exploit transformation patterns in unlabeled video to learn features that are useful for recognition.

Aside from inferring the transformation that implicitly separates a pair of training instances, another possibility is to explicitly predict the transformation parameters.  Recent work considers how the camera's ego-motion (e.g., as obtained from inertial sensors, GPS) can be exploited as supervision during CNN training~\cite{dinesh-ego-2015,agrawal-move-2015}.
These methods also lack the higher-order relationships we propose.  Furthermore, they require training data annotated with camera/ego-pose parameters, which prevents them from learning with  ``in the wild" videos (like YouTube) for which the camera was not instrumented with external sensors to record motor changes.  In contrast, our method is free to exploit arbitrary unlabeled video data.

\changes{
Several recent papers~\cite{burda2015importance,walker2015dense,goroshin2015learning} have trained unsupervised image representations targeting specific narrow tasks. \cite{burda2015importance} learn efficient generative codes to synthesize images, while~\cite{walker2015dense} learn features to predict pixel-level optical flow maps for video frames. Contemporary with an earlier version of our work~\cite{ourArxiv}, \cite{goroshin2015learning} proposed to learn features that vary linearly in time, for the specific task of extrapolating future video frames given a pair of past frames. They report qualitative results for toy video frame synthesis.  While our formulation also encourages collinearity in the feature space, our aim is to learn generally useful features from real videos without supervision, and we report results on natural image scene, object, and action recognition tasks.
}

\vspace{-0.05in}
\section{Approach}\label{sec:approach}

Given auxiliary raw unlabeled video, we wish to learn an embedding amenable to a supervised classification task.  We pose this as a feature learning problem in a convolutional neural network, where the hidden layers of the network are tuned not only with the backpropagation gradients from a classification loss, but also with gradients computed from the unlabeled video that exploit its temporal steadiness.

\vspace{-0.02in}
\subsection{Notation and framework overview}\label{sec:overview}

A \emph{supervised training dataset} $\mathcal{S}=\{(\bm{x}_i,\bm{y}_i)\}$ provides target class labels $\bm{y}_i \in \mathcal{Y}=\left[1,2,..,C\right]$ for images $\bm{x}_i\in\mathcal{X}$ (represented in pixel space). The \emph{unsupervised training dataset} $\mathcal{U}=\{\bm{x}_t\}$ consists of ordered video frames, where $\bm{x}_t$ is the video frame at time instant $t$.\footnote{For notational simplicity, we will describe our method assuming that the unsupervised training data is drawn from a single continuous video, but it is seamless to train instead with a batch of unlabeled video clips.}

Importantly, we do \emph{not} assume that the video $\mathcal{U}$ necessarily stems from the same categories or even the same domain as images in $\mathcal{S}$.  For example, in results we will demonstrate cases where $\mathcal{S}$ and $\mathcal{U}$ consist of natural scene images and autonomous vehicle video, respectively; or Web photos of human actions and YouTube video spanning dozens of distinct activities.  The idea is that training with diverse unlabeled video should allow the learner to recover fundamental cues about how objects move, how scenes evolve over time, how occlusions occur, how illumination varies, etc., independent of their specific semantic content.

The full image-pixels-to-class label classifier we learn will have the compositional form $\hat{y}_{\bm{\theta},W}=f_{W} \circ \mathbf{z}_{\bm{\theta}}(.),$ where $\mathbf{z}_{\bm\theta}: \mathcal{X}\rightarrow \mathcal{R}^D$ is a $D$-dimensional feature map operating on images in the pixel space, and $f_W:\mathcal{R}^D \rightarrow \mathcal{Y}$ takes as input the feature map $\mathbf{z}_{\bm{\theta}}(\bm{x})$, and outputs the class estimate.  We learn a linear classifier $f_W$ represented by a $C\times D$ weight matrix $W$ with rows $\bm{w}_1,\dots,\bm{w}_C$.  At test time, a novel image is classified as $\hat{y}_{\bm{\theta},W} = \argmax_i \bm{w}_i^T \mathbf{z}_{\bm{\theta}}(\bm{x})$.

To learn the classifier $\hat{y}_{\bm{\theta},W}$, we optimize an objective function of the form:
\begin{equation}
  (\bm{\theta}^*, W^*)=\argmin_{\bm{\theta},W} L_s(\bm{\theta},W,\mathcal{S})+\lambda L_u(\bm{\theta}, \mathcal{U}),
  \label{eq:framework}
\end{equation}
where $L_s(.)$ represents the supervised classification loss, $L_u(.)$ represents an unsupervised regularization loss term, and $\lambda$ is the regularization hyperparameter. The parameter vector $\bm{\theta}$ is common to both losses because they are both computed on the learned feature space $\mathbf{z}_{\bm{\theta}}(.)$. The supervised loss is a softmax loss:
\begin{equation}
L_s(\bm{\theta}, W,\mathcal{S}) = -\frac{1}{N_s} \sum_{i=1}^{N_s} \log(\sigma_{y_i}(W\mathbf{z}_{\bm{\theta}}(\bm{x}_i)),
  \label{sup_loss}
\end{equation}
where $\sigma_{y_i}(.)$ is the softmax probability of the correct class and $N_s$ is the number of labeled training instances in $\mathcal{S}$.

In the following, we first discuss how the unsupervised regularization loss $L_u(.)$ may be constructed to exploit temporal smoothness in video (\secref{sec:sfa}).  Then we generalize this to exploit temporal steadiness and other higher order coherence (\secref{sec:higher_sfa}). \secref{sec:siamese} then shows how a neural network corresponding to $\hat{y}_{\bm{\theta},W}$ may be trained to minimize \eqnref{eq:framework} above.

\subsection{Review: First-order temporal coherence}\label{sec:sfa}

As discussed above, slow feature analysis (SFA)~\cite{Wiskott2002} seeks to learn image features that vary slowly over the frames of a video, with the aim of learning useful invariances.  This idea of exploiting ``slowness'' or ``temporal coherence'' for feature learning has been explored in the context of neural networks
 ~\cite{Mobahi2009,drlim,bergstra-nips2009,zou2012deep,Goroshin2014}.  We briefly review that underlying objective before introducing the proposed higher order generalization of temporal coherence.

A temporal neighbor pair dataset $\mathcal{U}_2$ is first constructed from the unlabeled video $\mathcal{U}$, as follows:
\begin{align}
  \mathcal{U}_2=\{\langle(j, k), p_{jk}\rangle:& \bm{x}_j, \bm{x}_k \in \mathcal{U}\text{ and   }\nonumber\\
  &p_{jk}=\mathbbm{1}(0\leq j-k\leq T)\},
  \label{eq:pair_data}
\end{align}
where $T$ is the temporal neighborhood size, and the subscript 2 signifies that the set consists of \emph{pairs}. $\mathcal{U}_2$ indexes image pairs with neighbor-or-not binary annotations $p_{jk}$, automatically extracted from the video.  We discuss the setting of $T$ in results.  In general, one wants the time window spanned by $T$ to include motions that are small enough to be label-preserving, so that correct invariances are learned; in practice this is typically on the order of a second or less.

With this dataset, the SFA property translates as $\mathbf{z}_{\bm{\theta}}(\bm{x}_j) \approx \mathbf{z}_{\bm{\theta}}(\bm{x}_k), \forall p_{jk}=1$. A simple formulation of this as an unsupervised regularizing loss would be as follows:
\begin{equation}
  \vspace{-0.05in}
  R_2^\prime(\bm{\theta},\mathcal{U})=\sum_{(j,k)\in\mathcal{N}} d(\mathbf{z}_{\bm{\theta}}(\bm{x}_j),\mathbf{z}_{\bm{\theta}}(\bm{x}_k)),
  \label{eq:naive}
  \vspace{-0.05in}
\end{equation}
where $d(.,.)$ is a distance measure (e.g., $\ell_1$ in~\cite{Mobahi2009} and $\ell_2$ in~\cite{drlim}), and $\mathcal{N} \subset \mathcal{U}_2$ denotes the subset of ``positive" neighboring frame pairs \ie those for which $p_{jk} = 1$.   This loss by itself admits problematic minimizers such as $\mathbf{z}_{\bm{\theta}}(\bm{x})=0,{ }\forall \bm{x} \in \mathcal{X}$, which corresponds to $R_2^\prime=0$. Such solutions may be avoided by a \emph{contrastive}~\cite{drlim} version of the loss function that also exploits ``negative'' (non-neighbor) pairs:
\vspace{-0.05in}
\begin{align}
  &R_2(\bm{\theta},\mathcal{U}) =\sum_{(j,k)\in\mathcal{U}_2} D_\delta(\mathbf{z}_{\bm{\theta}}(\bm{x}_j),\mathbf{z}_{\bm{\theta}}(\bm{x}_k), p_{jk})\nonumber \\
  &=\sum_{(j,k)\in\mathcal{U}_2} p_{jk}~d(\mathbf{z}_{\bm{\theta}j},\mathbf{z}_{\bm{\theta}k})+\overline{p_{jk}}~\max(\delta-d(\mathbf{z}_{\bm{\theta}j},\mathbf{z}_{\bm{\theta}k}),0),
  \label{eq:drlim}
\end{align}
where $\mathbf{z}_{\bm{\theta}i}$ denotes $\mathbf{z}_{\bm{\theta}}(\bm{x}_i)$ and $\overline{p}=1-p$.
As shown above, the contrastive loss $D_{\delta}(\bm{a},\bm{b},p)$ penalizes distance  between $\bm{a}$ and $\bm{b}$ when the pair are neighbors ($p=1$), and encourages distance between them when they are not ($p=0$), up to a margin $\delta$.  %

\subsection{Higher-order temporal coherence}\label{sec:higher_sfa}

The slow feature formulation of~\eqnref{eq:drlim} encourages feature maps that produce small first-order temporal derivatives in the learned feature space: $d\mathbf{z}_{\bm{\theta}}(\bm{x}_t)/dt \approx 0$.  This first-order temporal coherence is restricted to learning to ignore small jitters in the visual signal.

Our idea is to model higher order temporal coherence in the unlabeled video, so that the features can further capture rich structure in \emph{how} the visual content changes over time.  In the general case, this means we want a regularizer that encourages higher order derivatives to be small:  $d^n\mathbf{z}_{\bm{\theta}}(\bm{x}_t)/dt^n \approx 0, \forall n=1,2,..N$.  Accordingly, we need to generalize from pairs of temporally close frames to tuples of frames.

In this work, we focus specifically on learning \emph{steady} features---the second-order case, which can be encoded with triplets of frames, as we will see next. In a nutshell, whereas slow learning insists that the features not change too quickly, steady learning insists that feature \emph{changes} in the immediate future remain similar to those in the recent past.

\begin{figure*}[ht]
  \centering
  \includegraphics[width=0.9\textwidth]{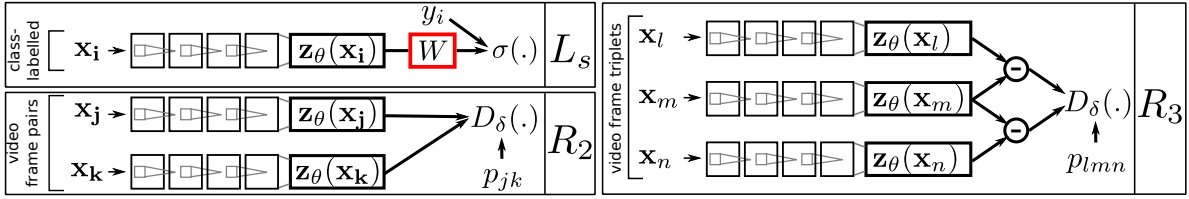}
  \caption{\small{``Siamese'' network configuration (shared weights for the $\mathbf{z}_{\bm\theta}$ layer stacks) with portions corresponding to the 3 terms $L_s$, $R_2$ and $R_3$ in our objective. $R_2$ and $R_3$ compose the unsupervised loss $L_u$ in~\eqnref{eq:framework}.  $L_s$ is the supervised loss for recognition in static images.}}
  \label{fig:network_config}
  \vspace{-0.15in}
\end{figure*}

First, we create a triplet dataset $\mathcal{U}_3$ from the unlabeled video $\mathcal{U}$ as:
\begin{align}
  \mathcal{U}_3=\{\langle(l, m, &n), p_{lmn}\rangle: \bm{x}_l, \bm{x}_m, \bm{x}_n \in \mathcal{U}\text{ and   }\nonumber\\
  & p_{lmn}=\mathbbm{1}( 0\leq m-l=n-m\leq T)\}.
  \label{eq:triplet_data}
\end{align}
$\mathcal{U}_3$ indexes image triplets with binary annotations indicating whether they are in-sequence, evenly spaced frames in the video, within a temporal neighborhood $T$.  In practice, we select ``negatives'' ($p_{lmn}=0$) from triplets where $m-l\leq T$ but $n-m\geq 2T$ to provide a buffer and avoid noisy negatives.

We construct our steady feature analysis regularizer using these triplets, as follows:
\vspace{-0.05in}
\begin{equation}
  R_3(\bm{\theta},\mathcal{U}) =\sum_{(l,m,n)\in\mathcal{U}_3} D_\delta(\mathbf{z}_{\bm{\theta}l}-\mathbf{z}_{\bm{\theta}m}, ~\mathbf{z}_{\bm{\theta}m}-\mathbf{z}_{\bm{\theta}n},~p_{lmn}),
  \label{eq:triplim}
\vspace{-0.05in}
\end{equation}
where $\mathbf{z}_{\bm{\theta}l}$ is again shorthand for $\mathbf{z}_{\bm{\theta}}(\bm{x}_l)$ and $D_\delta$ refers to the contrastive loss defined above. For positive triplets---meaning those occurring in sequence and within a temporal neighborhood---the above loss penalizes distance between the adjacent pairwise feature \emph{difference} vectors.  For negative triplets, it \emph{encourages} this distance, up to a maximum margin distance $\delta$. Effectively, $R_3$ encourages the feature representations of positive triplets to be collinear \ie $\mathbf{z}_{\bm{\theta}}(\bm{x}_l)-\mathbf{z}_{\bm{\theta}}(\bm{x}_m)\approx \mathbf{z}_{\bm{\theta}}(\bm{x}_m)-\mathbf{z}_{\bm{\theta}}(\bm{x}_n)$.  See Figure~\ref{fig:concept}.

Our final optimization objective combines the first and second order losses (\eqnref{eq:drlim} and \eqref{eq:triplim}) into the unsupervised regularization term:
\vspace{-0.05in}
\begin{equation}
L_u(\bm{\theta},\mathcal{U})=R_2(\bm{\theta},\mathcal{U})+\lambda^\prime R_3(\bm{\theta},\mathcal{U}),
\label{eq:final_reg}
\vspace{-0.05in}
\end{equation}
where $\lambda^\prime$ controls the relative impact of the two terms.  Recall this regularizer accompanies the classification loss in the main objective of \eqnref{eq:framework}.

\vspace*{-0.15in}
\paragraph{Beyond second-order coherence:}

The proposed framework generalizes naturally to the $n$-th order, by defining $R_n$ analogously to~\eqnref{eq:triplim} using a contrastive loss over $(n-1)$-th order discrete derivatives, computed over recursive differences on $n$-tuples.
While in principle higher $n$ would more thoroughly exploit patterns in video, there are potential practical drawbacks. As $n$ grows, the number of samples $|\mathcal{U}_n|$ would likely need to also grow to cover the space of $n$-frame motion patterns, \KG{requiring more training time, compute power, and memory.}  Besides, discrete $n$-th derivatives computed over large $n$-frame time windows may grow less reliable, \KG{assuming steadiness degrades over longer temporal windows in typical visual phenomena}.  Given these considerations, we focus on  second-order steadiness combined with slowness, and find that slow and steady does indeed win the race (Sec~\ref{sec:exp}).
The empirical question of applying $n>2$ is left for future work.

\vspace*{-0.15in}
\paragraph{Equivariance-inducing property of $R_3(\bm{\theta},\mathcal{U})$:}

While first-order coherence encourages invariance, the proposed second-order coherence may be seen as encouraging the more general property of \emph{equivariance}.
$\mathbf{z}(.)$
is equivariant to an image transformation $g$ if there  exists some ``simple'' function $\mathbf{f}_g:\mathcal{R}^D\rightarrow\mathcal{R}^D$ such that $\mathbf{z}(g\bm{x})\approx \mathbf{f}_g( \mathbf{z}(\bm{x})).$ Equivariance has been found to be useful for visual representations~\cite{hinton2011,schmidt2012learning,vedaldi2014,dinesh-ego-2015}. To see how feature steadiness is related to equivariance, consider a video with frames $\bm{x}_t, 1\leq t \leq T$. Given a small temporal neighborhood $\Delta t$, frames $\bm{x}_{t+\Delta t}$ and $\bm{x}_{t}$ must be related by a \emph{small} transformation $g$ (small because of \emph{first} order temporal coherence assumption) \ie $\bm{x}_{t+\Delta t}=g\bm{x}_t$. Assuming \emph{second} order coherence of video, this transformation $g$ itself remains approximately constant in a small temporal neighborhood, so that, in particular, $\bm{x}_{t+2\Delta t}\approx g\bm{x}_{t+\Delta t}$.

  Now, for equivariant features $\mathbf{z}(.)$, by the definition of equivariance and the observations above, $\mathbf{z}(\bm{x}_{t+2\Delta t})\approx \mathbf{f}_g(\mathbf{z}(\bm{x}_{t+\Delta t}))\approx\mathbf{f}_g \circ \mathbf{f}_g(\mathbf{z}(\bm{x}_{t}))$. Further, given that $g$ is a small transformation, $\mathbf{f}_g$ is well-approximated in a small neighborhood by its first order Taylor approximation, so that: (1) $\mathbf{z}(\bm{x}_{t+\Delta t})\approx \mathbf{z}(\bm{x}_{t})+\mathbf{c}(t)$, and (2) $\mathbf{z}(\bm{x}_{t+2 \Delta t})\approx \mathbf{z}(\bm{x}_{t})+2\mathbf{c}(t)$. In other words, under the realistic assumption that natural videos evolve smoothly, within small temporal neighborhoods, feature equivariance is equivalent to the second order temporal coherence formulated in Eq~\eqref{eq:triplim}, with $l,m,n$ set to $t, t+\Delta t, t+ 2\Delta t$ respectively.
This connection between equivariance and the second order temporal coherence induced by $R_3$ helps motivate why we can expect our feature learning scheme to benefit recognition.

\vspace{-0.05in}
\subsection{Neural networks for the feature maps}\label{sec:siamese}

We use a convolutional neural network (CNN) architecture to represent the feature mapping function $\mathbf{z}_{\bm{\theta}}(.)$. The parameter vector $\bm{\theta}$ represents the CNN's learned layer weight matrices. See Sec~\ref{sec:exp_setup} and Supp for architecture choices.

To optimize \eqnref{eq:framework} with the regularizer in \eqnref{eq:final_reg}, we employ standard mini-batch stochastic gradient descent (as implemented in~\cite{caffe}) in a ``Siamese'' setup, with 6 replicas of the stack $\mathbf{z}_{\bm{\theta}}(.)$, as shown in Fig~\ref{fig:network_config}, 1 stack for $L_s$ (input: supervised training samples $\bm{x}_i$), 2 for $R_2$ (input: temporal neighbor pairs $(\bm{x}_j, \bm{x}_k)$) and 3 for $R_3$ (input: triplets $(\bm{x}_l, \bm{x}_m, \bm{x}_n)$).
The shared layers are initialized to the same random values and modified by the same gradients (sum of the gradients of the 3 terms) in each training iteration, so they remain identical throughout. See Supp for details. %

\begin{table*}[t]
  \centering
  \ra{1.5}
  \resizebox{1\textwidth}{!}{%
  \begin{tabular}{@{}lc|ccll|lll@{}}
    \toprule
    Task                       & Img/frame dims                  & \#Classes & Recog. Task & \#Train & \#Test & Unsup. Input Type & \#Pairs (1:3) & \#Triplets (1:1)\\
    \midrule
    NORB$\rightarrow$NORB      & 96$\times$96$\times$1 & 25  & object & 150     & 8100   & pose-reg. images  & 50,000      & 75,000\\
    KITTI$\rightarrow$SUN      & 32$\times$32$\times$1 & 397 & scene  & 2382    & 7940   & car-mounted video & 100,000     & 100,000\\
    HMDB$\rightarrow$PASCAL-10 & 32$\times$32$\times$3 & 10  & action & 50       & 2000   & web video         & 100,000     & 100,000\\
    \bottomrule
  \end{tabular}
  \quad
  \quad
  \begin{tabular}{@{}l|ccc}
  \toprule
  Datasets$\rightarrow$            & NORB          & KITTI         & HMDB          \\
  \midrule
  \textsc{sfa-1}~\cite{Mobahi2009} & 0.95          & 31.04         & 2.70          \\
  \textsc{sfa-2}~\cite{drlim}      & 0.91          & 8.39          & 2.27          \\
  \textsc{ssfa} (ours)             & \textbf{0.53} & \textbf{7.79} & \textbf{1.78} \\
  \bottomrule
  \end{tabular}
}
\caption{\small{\textbf{Left:} Statistics for the unsupervised and supervised datasets ($\mathcal{U} \rightarrow \mathcal{S}$) used in the recognition tasks (positive to negative ratios for pairs and triplets indicated in headers). \textbf{Right:} Sequence completion normalized correct candidate rank $\eta$. Lower is better. (See \secref{sec:sanity_check}.)}}
  \label{tab:task_stats}
  \label{tab:seq_completion}
  \vspace{-0.15in}
\end{table*}

\vspace{-0.1in}
\section{Experiments}\label{sec:exp}

We test our approach using five challenging public datasets for three tasks---object, scene, and action recognition---spanning 432 categories. We also analyze its ability to learn higher order temporal coherence with a sequence completion task.

\vspace{-0.05in}
\subsection{Experimental setup}\label{sec:exp_setup}

Our three recognition tasks (specified by the names of the unsupervised and supervised datasets as $\mathcal{U}\rightarrow\mathcal{S}$) are NORB$\rightarrow$NORB object recognition, KITTI$\rightarrow$SUN scene recognition and HMDB$\rightarrow$PASCAL-10 single-image action recognition. \tabref{tab:task_stats} (left) summarizes  key dataset statistics.  %

\vspace*{-0.15in}
\paragraph{Supervised datasets $\mathcal{S}$:}

(1) \textbf{NORB}~\cite{norb} has 972 images each of 25 toys against clean backgrounds captured over a grid of camera elevations and azimuths.   (2) \textbf{SUN}~\cite{sun} contains Web images of 397 scene categories. (3) \textbf{PASCAL-10}~\cite{pascal} is a still-image human action recognition dataset with 10 categories.  For all three datasets, we use few labeled training images (see \tabref{tab:task_stats}), since unsupervised regularization schemes should have most impact when labeled data is scarce~\cite{dinesh-ego-2015, Mobahi2009}.  This is an important scenario, given the ``long tail" of categories lacking ample labeled exemplars.

\vspace*{-0.15in}
\paragraph{Unsupervised datasets $\mathcal{U}$:}

(1) \textbf{NORB} consists of pose-registered turntable images (not video), but it is straightforward to generate the pairs and triplets for $\mathcal{U}_2$ and $\mathcal{U}_3$ assuming smooth motions in the annotated pose space.  We mine these pairs and triplets from among the 648 images per class that are not used for testing.
(2) \textbf{KITTI}~\cite{kitti} has videos captured from a car-mounted camera in a variety of locations around the city of Karlsruhe. Scenes are largely static except for traffic, but there is large and systematic camera motion. %
(3) \textbf{HMDB}~\cite{hmdb} contains 6849 short Web and movie video clips %
 containing 51 diverse actions. %
 We select 1000 clips at random. While some videos include camera motion (\eg to follow an athlete running), most have stationary cameras and small human pose-change motions.
 The time window $T$ is a hyperparameter of both our method as well as existing SFA methods. We fix $T=2$ and $T=0.5$ seconds for KITTI and HMDB, respectively, based on cross-validation for best performance by the SFA baselines.

\vspace*{-0.15in}
 \paragraph{Baselines:} We compare our slow-and-steady feature analysis approach (\textsc{ssfa}) to four methods, including two key existing methods for learning from unlabeled video.  The three unsupervised baselines are: (1) \textsc{unreg}: An unregularized network trained only on the supervised training samples $\mathcal{S}$. (2) \textsc{sfa-1}: An SFA approach proposed in~\cite{Mobahi2009} that uses $\ell_1$ for $d(.)$ in Eq~\ref{eq:drlim}. (3) \textsc{sfa-2}: Another SFA variant~\cite{drlim} that sets the distance function $d(.)$ to the $\ell_2$ distance in Eq~\ref{eq:drlim}.
 The \textsc{sfa} methods train with the unlabeled pairs, while \textsc{ssfa} trains with both the pairs and triplets.

 These comparisons are most crucial to gauge the impact of the proposed approach versus the state of the art for feature learning with unlabeled video.  However, we are also interested to what extent learning from unlabeled video can even start to compete with methods learned from heavily labeled data (which costs substantial human effort).  Thus, we also compare against a \emph{supervised} pretraining and finetuning approach denoted \textsc{sup-ft} (details in \secref{sec:recog_result}).

\vspace*{-0.1in}
\paragraph{Network architectures:}

For the NORB$\rightarrow$NORB task, we use a fully connected network architecture: input $\rightarrow$ 25 hidden units $\rightarrow$ ReLU nonlinearity $\rightarrow$  $D$=25 features. For the other two tasks, we resize images to $32 \times 32$ to allow fast and thorough experimentation with standard CNN architectures known to work well with tiny images~\cite{cuda-convnet}, producing $D$=64-dimensional features. Recognition tasks on 32$\times$32 images are much harder than with full-sized images, so these are highly challenging tasks. All networks are optimized with Nesterov-accelerated stochastic gradient descent until validation classification loss converges or begins to increase. Optimization hyperparameters are selected greedily through cross-validation in the following order: base learning rate, $\lambda$ and $\lambda^\prime$ (starting from $\lambda$=$\lambda^\prime$=0). The relative scales of the margin parameters $\delta$ of the contrastive loss $D_{\delta}(.)$ in \eqnref{eq:drlim} and \eqnref{eq:triplim} are validated per dataset. See Supp for more details on the 32$\times$32 architecture, data pre-processing and optimization.

\begin{figure*}[t]
\centering
\includegraphics[width=1\textwidth]{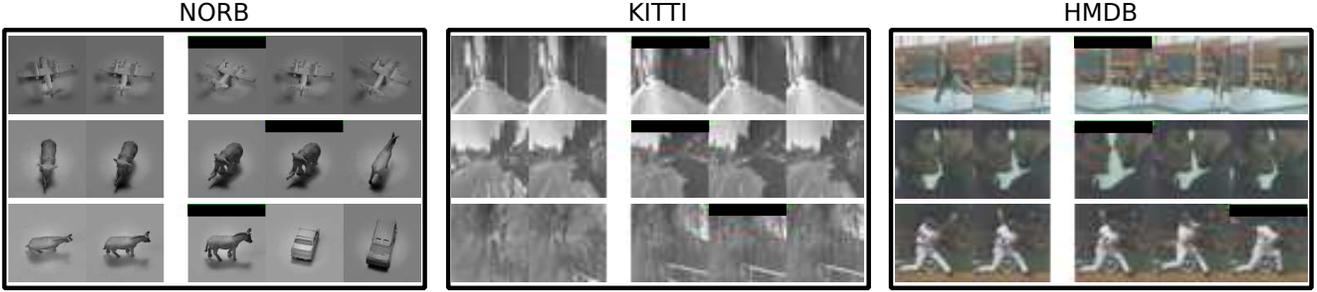}
\caption{\small{Sequence completion examples from all three video datasets. In each instance, a query pair is presented on the left, and the top three completion candidates as ranked by our method are presented on the right. Ground truth frames are marked with black highlights.}}
\label{fig:qual}
\vspace{-0.15in}
\end{figure*}

\subsection{Quantifying steadiness}\label{sec:sanity_check}

First we use a sequence completion task to analyze how well the desired steadiness property is induced in the learned features.
We compose a set of sequential triplets from the pool of test images, formed similarly to the positives in \eqnref{eq:triplet_data}.  At test time, given the first two images of each triplet, the task is to predict what the third looks like.

We apply our \textsc{ssfa} to infer the missing triplet item as follows.  Recall that our formulation encourages sequential triplets to be collinear in the feature space. As a result, given $\mathbf{z}_{\bm{\theta}}(\bm{x}_1)$ and $\mathbf{z}_{\bm{\theta}}(\bm{x}_2)$, we can extrapolate $\mathbf{z}_{\bm{\theta}}(\bm{x}_3)$ as $\tilde{\mathbf{z}}_{\bm{\theta}}(\bm{x}_3)=2\mathbf{z}_{\bm{\theta}}(\bm{x}_2)-\mathbf{z}_{\bm{\theta}}(\bm{x}_1)$. To backproject to the image space, we identify an image closest to $\tilde{\mathbf{z}}_{\bm{\theta}}(\bm{x}_3)$ in feature space.  Specifically, we take a large pool $\mathcal{C}$ of candidate images, map them all to their features via $\mathbf{z}_{\bm{\theta}}$, and rank them in increasing order of distance from $\tilde{\mathbf{z}}_{\bm{\theta}}(\bm{x}_3)$.   The rank $r$ of the correct candidate $\bm{x}_3$ is now a measure of sequence completion performance. See Supp for details. %

 Tab~\ref{tab:seq_completion} (right) reports the mean percentile rank $\eta=\mathbb{E}[{r}/{|\mathcal{C}|}]\times 100$ over all query pairs. Lower $\eta$ is better.
Clearly, our \textsc{ssfa} regularization induces steadiness in the feature space, reducing $\eta$ nearly by half compared to baseline regularizers on NORB and by large margins on HMDB too. Our regularizer $R_3$ is closely matched to this task, so these gains are expected. Note however that these gains are reported after training to minimize the \emph{joint} objective, which includes $L_s$ and $R_2$, apart from $R_3$, and with regularization weights tuned for \emph{recognition} tasks.

\figref{fig:qual} shows sequence completion examples from all 3 video datasets. Particularly impressive results are the third NORB example (where despite a difficult viewpoint, the sequence is completed correctly by the top-ranked candidate), and the third HMDB example, where a highly dynamic baseball pitch sequence is correctly completed by the third ranked image. The top-ranked candidate for this example illustrates a common failure mode---the second image of the query pair is itself picked to complete the sequence. This may reflect the fact that HMDB sequences in particular exhibit very little motion (camera motions rare, mostly small object motions). Usually, as in the third KITTI example, even the top-ranked candidates other than the ground truth frame are highly plausible completions.

\subsection{Recognition results}\label{sec:recog_result}

\paragraph{Unlabeled video as a prior for supervised recognition:}
Now we report results on the 3 unsupervised-to-supervised recognition tasks. %
\tabref{tab:recog_result} shows the results. Our \textsc{ssfa} method comprehensively outperforms not only the purely supervised \textsc{unreg} baseline, but also the popular \textsc{sfa-1} and \textsc{sfa-2} slow feature learning approaches, beating the best baseline for each task by 9\%, 36\% and 9\% respectively.  The results on KITTI$\rightarrow$SUN and HMDB$\rightarrow$PASCAL-10 are particularly impressive because the unsupervised and supervised dataset domains are mismatched. All KITTI data comes from a single car-mounted road-facing camera driving through the streets of one city, whereas SUN images are downloaded from the Web, captured by different cameras from diverse viewpoints, and cover 397 scene categories mostly unrelated to roads. PASCAL-10 images are bounding-box-cropped and therefore centered on single persons, while HMDB videos, which are mainly clips from movies and Web videos, often feature multiple people, are not as tightly focused on the person performing the action, and are of low quality, sometimes with overlaid text \etc

\begin{table}
  \centering
  \ra{1.5}
  \resizebox{1\linewidth}{!}{%
      \begin{tabular}{@{}l|cccc@{}}
    \toprule
    Task type$\rightarrow$              & Objects & \multicolumn{2}{c}{Scenes} & Actions \\ \midrule
    Datasets$\rightarrow$               & NORB$\rightarrow$NORB   & \multicolumn{2}{c}{KITTI$\rightarrow$SUN} & HMDB$\rightarrow$PASCAL-10                                        \\
    Methods$\downarrow$                 & [25 cls]                & [397 cls]                                 & [397 cls, top-10]                & [10 cls]                    \\
    \midrule
    random                              & 4.00                    & 0.25                                      & 2.52                             & 10.00                          \\
    \textsc{unreg}                      & 24.64$\pm$0.85          & 0.70$\pm$0.12                             & 6.10$\pm$0.67                    & 15.34$\pm$0.28             \\
    \textsc{sfa-1}~\cite{Mobahi2009}    & 37.57$\pm$0.85          & 1.21$\pm$0.14                             & 8.24$\pm$0.25                    & 19.26$\pm$0.45             \\
    \textsc{sfa-2}~\cite{drlim}         & 39.23$\pm$0.94          & 1.02$\pm$0.12                             & 6.78$\pm$0.32                    & 19.04$\pm$0.24             \\
    \textsc{ssfa} (ours)                & \textbf{42.83$\pm$0.33} & \textbf{1.65$\pm$0.04}                    & \textbf{9.19$\pm$0.10}           & \textbf{20.95$\pm$0.13} \\
    \bottomrule
    \end{tabular}
  }
  \caption{Recognition results (mean $\pm$ standard error of accuracy \% over 5 repetitions) (\secref{sec:recog_result}).  Our method outperforms both existing slow feature/temporal coherence methods and the unregularized baseline substantially, across three distinct recognition tasks.}
\label{tab:recog_result}
\vspace{-0.18in}
\end{table}

Aside from the diversity of tasks (object, scene, and action recognition), our unsupervised datasets also exhibit diverse types of motion. NORB is generated from planned, discrete camera manipulations around a central object of interest. The KITTI camera moves through a real largely static landscape in smooth motions on roads at varying speeds. HMDB videos on the other hand are usually captured from stationary cameras with a mix of large and small foreground and background object motions. Even the dynamic camera videos in HMDB are sometimes captured from hand-held devices leading to jerky motions, where our temporal steadiness assumptions might be stressed.

\vspace*{-0.16in}
\paragraph{Pairing unsupervised and supervised datasets:}

Thus far, our pairings of unsupervised and supervised datasets reflect our attempt to learn from video that \emph{a priori} seems related to the ultimate recognition task, \eg HMDB human action videos are paired with PASCAL-10 Action still images.  However, as discussed above, the domains are only roughly aligned. Curious about the impact of the choice of unlabeled video data, we next try swapping out HMDB for KITTI in the PASCAL action recognition task. On this new KITTI$\rightarrow$PASCAL task, we still easily outperform our nearest baseline, although our gain drops by $\approx$ 0.9\% (\textsc{sfa-2}:19.06\% vs. our \textsc{ssfa}:20.01\%).
Despite the fact that the human motion dynamics of HMDB ostensibly match the action recognition task better than the egomotion dynamics of KITTI (where barely any people are visible), we maintain our advantage over the purely slow methods.  This indicates that there is reasonable flexibility in the choice of unlabeled videos fed to \textsc{ssfa}.

\vspace*{-0.16in}
\paragraph{Increasing supervised training sets:}

\changes{Thus far, we have kept labeled sets small to simulate the ``long tail'' of categories with scarce training samples where priors \cc{like ours and the baselines'} have most impact. In a preliminary study for larger training pools, we now increase SUN training set sizes from 6 to 20 samples per class for KITTI$\rightarrow$SUN.
Our method retains a 20\% gain over existing slow methods (\textsc{ssfa}: 3.24\% vs \textsc{sfa-2}: 2.65\%).  This suggests our approach is valuable even with larger supervised training sets.}

\vspace*{-0.16in}
\paragraph{Varying unsupervised training set size:} \changes{To observe the effect of unsupervised training set size, we now restrict \textsc{ssfa} to use varying-sized subsets of unlabeled video on the HMDB$\rightarrow$PASCAL-10 task. Performance scales roughly log-linearly with the duration of video observed,\footnote{At 3, 12.5, 25, and 100\% resply. of the full unlabeled dataset ($\approx$32k frames), performance is 18.06, 19.74, 20.36, and 20.95\% (see Supp)} suggesting that even larger gains may be achieved simply by training \textsc{ssfa} with more freely available unlabeled video.}

\vspace*{-0.16in}
\paragraph{Purely unsupervised feature learning:}  \changes{We now evaluate the usefulness of features trained to optimize the unsupervised \textsc{ssfa} loss $L_u$ (Eq~\eqref{eq:final_reg}) alone. Features trained on HMDB are evaluated at various stages of training, on the task of $k$-nearest neighbor classification on PASCAL-10 ($k=$5, and 100 training images per action). Starting at $\approx$ 17.8\% classification accuracy for randomly initialized networks, unsupervised \textsc{ssfa} training steadily improves the discriminative ability of features to 19.62, 20.32 and 22.14\% after 1, 2 and 3 passes respectively over training data (see Supp). This shows that \textsc{ssfa} can train useful image representations even without jointly optimizing a supervised objective.
}

\vspace*{-0.16in}
\paragraph{Comparison to supervised pretraining and finetuning:} Recently, a two-stage supervised pretraining and finetuning strategy (\textsc{sup-ft}) has emerged as the leading approach to solve visual recognition problems with limited training data where high-capacity models like deep neural networks may not be directly learned\changes{~\cite{rcnn,decaf,oquab2014learning,style}}. In the first stage (``supervised pretraining''), a neural network ``NET1'' is first trained on a related problem for which large training datasets \emph{are} available. In a second stage (``finetuning''), the weights from NET1 are used to initialize a second network (``NET2'') with similar architecture.  NET2 is then trained on the target task, using reduced learning rates to minimally modify the features learned in NET1.%

In principle, completely unsupervised feature learning approaches like ours have important advantages over the \textsc{sup-ft} paradigm. In particular, (1) they can leverage essentially infinite unlabeled data without requiring expensive human labeling effort thus potentially allowing the learning of higher capacity models and (2) they do not require the existence of large \emph{``related"} supervised datasets from which features may be meaningfully transferred to the target task. While the pursuit of these advantages continues to drive vigorous research, unsupervised feature learning methods still underperform supervised pretraining for image classification tasks, where great effort has gone into curating large labeled databases, e.g., ImageNet~\cite{deng2009imagenet}, CIFAR~\cite{cifar100}.

\begin{figure}[]
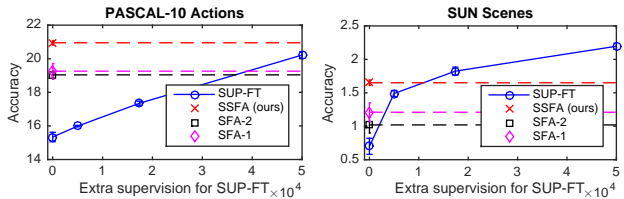

   \centering
\subfloat{
      \includegraphics[width=0.48\linewidth]{figs/PASCAL_sup_ft_plots}
}
\subfloat{
      \includegraphics[width=0.48\linewidth]{figs/SUN_sup_ft_plots}
    }
  \caption{Comparison to CIFAR-100 supervised pretraining \textsc{sup-ft}, at various supervised training set sizes.  Flat dashed lines reflect that our method (and SFA) always use zero additional labels.}
  \label{fig:sup_ft_comparison}
  \vspace{-0.15in}
\end{figure}

As a final experiment, we examine how the proposed unsupervised feature learning idea competes with the popular supervised pretraining model.  To this end, we adopt the CIFAR-100 dataset consisting of 100 diverse object categories as a basis for supervised pretraining.\footnote{We choose CIFAR-100 for its compatibility with the 32 $\times$ 32 images used throughout our results, which let us leverage standard CNN architectures known to work well with tiny images~\cite{cuda-convnet}.}  The new baseline \textsc{sup-ft} trains NET1 on CIFAR (see Supp), then finetunes NET2 for either PASCAL-10 action or SUN scene recognition tasks using the exact same (few) labeled instances given to our method.  In parallel, our method ``pretrains" only via the SSFA regularizer learned with unlabeled HMDB / KITTI video respectively for the two tasks.  Our method uses \emph{zero} labeled CIFAR data.

Fig ~\ref{fig:sup_ft_comparison} shows the results.
On PASCAL-10 action recognition (left), our method significantly outperforms \textsc{sup-ft} pretrained with all 50,000 images of CIFAR-100! Gathering image labels from the crowd for large multi-way problems can take on average 1 minute per image~\cite{olga-ijcv2015}, meaning we are getting better results while also saving $\sim$ 830 hours of human effort. On SUN scene recognition (right), \textsc{ssfa} outperforms \textsc{sup-ft} with 5K labels and remains competitive even when the supervised method has a 17,500 label advantage. However, \textsc{sup-ft-50k}'s advantage on the SUN task is more noticeable; its gain is similar to our gain over the best slow-feature method.

The upward trend in accuracy for \textsc{sup-ft} with more CIFAR-100 labeled data indicates that it successfully transfers generic recognition cues to the new tasks. On the other hand, the fact that it fares worse on PASCAL actions than SUN scenes reinforces that \emph{supervised} transfer depends on having large curated datasets in a \emph{strongly related} domain.  In contrast, our approach successfully ``transfers" what it learns from purely unlabeled video.
In short, our method can achieve better results with substantially less supervision.  More generally, we view it as an exciting step towards unlabeled video bridging the gap between unsupervised and supervised pretraining for visual recognition.

\vspace{-0.05in}
\section{Conclusion}

\changes{
  We formulated an unsupervised feature learning approach that exploits higher order temporal coherence in unlabeled video, and demonstrated its powerful impact for several recognition tasks. Despite over 15 years of research surrounding slow feature analysis (SFA), its variants and applications, to the best of our knowledge, we are the first to identify that SFA is only the first order approximation of a more general temporal coherence idea. This basic observation leads to our intuitive approach that can be easily plugged into applications where first order temporal coherence has already been found useful~\cite{Mobahi2009,bergstra-nips2009,zou2012deep,Goroshin2014,wang-gupta-2015,drlim,zhang-pami2012,zafeiriou-iccv2013,nater-bmvc2011,liwicki-accv2012}. To our knowledge, ours are the first results where unsupervised learning from video actually surpasses the accuracy of today's favored approach, heavily supervised pretraining.}

\vspace{0.04in}
\small{\noindent \textbf{Acknowledgements:} We thank Texas Advanced Computing Center for their generous support. This work was supported in part by ONR YIP N00014-15-1-2291.}
\vspace{-0.1in}

\clearpage

\pagebreak
\small{
\bibliographystyle{plain}
\bibliography{refs,jd_refs}
}

\clearpage

\section{Appendix}
\begin{figure}[]
  \centering
  \includegraphics{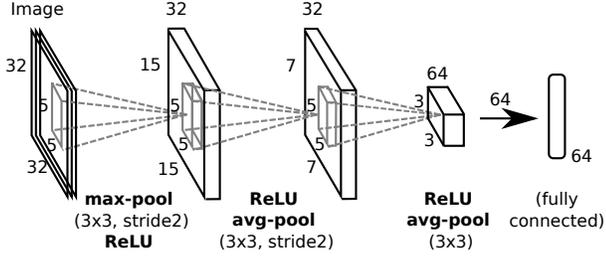}
  \caption{32$\times$32 CNN architecture used for the KITTI$\rightarrow$SUN and HMDB$\rightarrow$PASCAL-10 tasks}
  \label{fig:kitti_cnn}
\end{figure}

We now provide supplementary details on (1) the CNN architecture used in our SUN and PASCAL-10 experiments, (2) the sequence completion task used to quantify steadiness, (3) our experiments with varying sizes of unsupervised training datasets, (4) our experiments with purely unsupervised feature learning, (5) pre-processing steps for the datasets used in our experiments, (6) optimization-related details, and (7) details of the supervised pretraining and finetuning baseline \textsc{sup-ft} from the paper. We also show samples of all the real image datasets used in our experiments.
\paragraph{32$\times$32 images CNN architecture:}
The 32$\times$32 CNN architecture~\cite{cuda-convnet} representing $\mathbf{z}_{\bm\theta}$, used for the KITTI$\rightarrow$SUN and HMDB$\rightarrow$PASCAL-10 tasks is shown in Fig~\ref{fig:kitti_cnn}.

\vspace{-0.1in}
\paragraph{Quantifying steadiness - details}
As described in the main paper (Sec 4.2), the candidate set $\mathcal{C}$ for NORB was straightforward to construct -- the entire NORB test image set was used. For the video datasets KITTI and HMDB though, it would have been practically difficult to include all image frames in the candidate set $\mathcal{C}$. To avoid having to compute features and perform nearest neighbor search over too large a number of frames, we formed a randomly sub-sampled $\mathcal{C}$ instead, as follows. Starting from empty $\mathcal{C}$, we added (1) all the unique images among the query pairs (2) their corresponding ground truth completion images and (3) a minimum number $N$ of randomly chosen frames from each video represented within $\mathcal{C}$ until this point. This ensures that the task is non-trivial by adding distractors from the same video as the ground truth candidate image, which are likely to have similar appearance. We used $N$=10 for KITTI and $N$=5 for HMDB to keep the total numbers of images manageable. Finally, we select from $|\mathcal{C}|=$8100, 5000 and 5000 candidates respectively for NORB, KITTI and HMDB, for each of $N=$20,000, 1000 and 1,000 query pairs respectively for the three datasets.

\begin{figure}[h]
\centering
\includegraphics[width=0.7\linewidth]{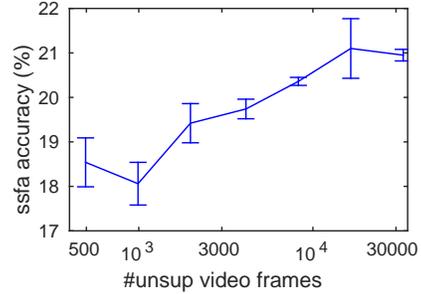}
\caption{\textsc{ssfa} classification accuracy vs.~duration of unsupervised video (mean, standard error over 5 runs).}
\label{fig:vary_unsup_size}
\end{figure}

\vspace{-0.1in}
\paragraph{Varying unsupervised training set size:} \changes{To observe the effect of unsupervised training set size, we now restrict \textsc{ssfa} to use varying-sized subsets of unlabeled video on the HMDB$\rightarrow$PASCAL-10 task. The full HMDB dataset has approximately 1000 videos, for a total of $\approx$32000 frames. Performance scales roughly log-linearly with the duration of video observed as shown in Fig~\ref{fig:vary_unsup_size}, suggesting that even larger gains may be achieved simply by training \textsc{ssfa} with more freely available unlabeled video.}

\begin{figure}[h]
\centering
\includegraphics[width=0.7\linewidth]{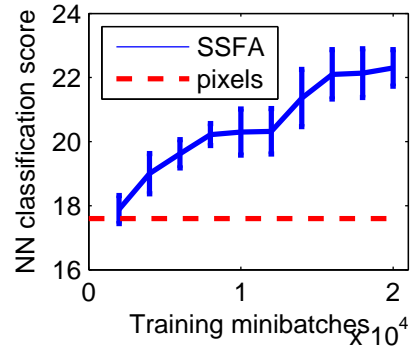}
\caption{\textsc{ssfa} k-NN accuracy improvement with \textsc{ssfa} training (mean, standard error over 5 runs).}
\label{fig:knn}
\end{figure}

\vspace{-0.1in}
\paragraph{Purely unsupervised feature learning:}  \changes{We evaluate the usefulness of features trained to optimize the unsupervised \textsc{ssfa} loss $L_u$ (main paper Eq~\eqref{eq:final_reg}) alone. Features trained on HMDB are evaluated at various stages of training, on the task of $k$-nearest neighbor classification on PASCAL-10 ($k=$5, and 100 training images per action). Fig~\ref{fig:knn} shows the results. Starting at $\approx$ 17.8\% classification accuracy for randomly initialized networks, unsupervised \textsc{ssfa} training steadily improves the discriminative ability of features. This shows that \textsc{ssfa} can train useful image representations even without jointly optimizing a supervised objective.
}

\vspace{-0.1in}
\paragraph{Dataset pre-processing details}
For all tasks, images are mean-subtracted and contrast-normalized before passing to the neural networks. In addition, for KITTI$\rightarrow$SUN, full KITTI frames were resized to 32$\times$32 and SUN images were cropped to KITTI aspect ratio before resizing to the same dimensions. Grayscale images were used in this task. Similarly, for HMDB$\rightarrow$PASCAL-10, HMDB frames were cropped to centered squares, and PASCAL-10 bounding boxes were expanded to the closest square before resizing to 32$\times$32. Resizing for KITTI$\rightarrow$SUN and HMDB$\rightarrow$PASCAL-10 was done to allow fast and thorough experimentation with standard CNN architectures known to work well with tiny images~\cite{cuda-convnet}.
On the SUN dataset apart from resizing, where we also lose information due to KITTI-aspect-ratio cropping, we verified that our baselines were legitimate by running a simple nearest neighbor baseline in the pixel space (standard approach for tiny images). This achieved 0.61\% accuracy compared to \textsc{UNREG}'s 0.70\%, given the same training data.

\vspace{-0.1in}
\paragraph{Optimization details}
We initialized according to the scheme proposed in~\cite{xavier}, and run Nesterov accelerated stochastic gradient descent using the open source Caffe~\cite{caffe} package. The base learning rate and regularization $\lambda$s are selected with greedy cross-validation.\footnote{our validated ($\lambda$,$\lambda^\prime$) values for NORB$\rightarrow$NORB, KITTI$\rightarrow$SUN, and HMDB$\rightarrow$PASCAL respectively are (0.1,0.3),(3,0.1), and (0.3,1)}
Specifically, for each task, the optimal base learning rate (from 0.1, 0.01, 0.001, 0.0001) was first identified for \textsc{unreg}. Next $\lambda$ was set through a logarithmic grid search (steps of $10^{0.5}$), with $\lambda^\prime$ set to 0 \ie this parameter was optimized for \textsc{sfa-2}. The margin parameter $\delta$ of the contrastive loss in $R_2(.)$ was set to 1.0 for all methods -- this affects the objective function only up to a feature scaling operation, and so may be set to any positive value. For \textsc{ssfa}, a similar search was then performed over $\lambda^\prime$ (logarithmic grid search with steps of $10^{0.5}$), and then a small search for the contrastive loss margin $\delta$ in $R_3(.)$ (over 0, 0.1 and 1).  Setting the margin to $\delta=0$ in a contrastive loss reduces it to the simple distance loss over positive samples.

On a single Tesla K-40 GPU machine, NORB$\rightarrow$NORB training tasks took $\approx$30 minutes, KITTI$\rightarrow$SUN tasks took $\approx$ 90 minutes, and HMDB$\rightarrow$PASCAL-10 tasks took $\approx$60 minutes. \textsc{ssfa} training took about 2x training time and 1.5x training epochs to converge, compared to \textsc{sfa} baselines, because of the more complex loss function.

\vspace{-0.1in}
\paragraph{Supervised pretraining and finetuning - details}
For the supervised pretraining and finetuning comparison experiments in Sec 4.3, we used the same neural network architecture as used for our approach and other baselines on the SUN scene and PASCAL-10 action recognition tasks (architecture shown in Fig~\ref{fig:kitti_cnn}). A 100-way softmax classifier was trained on the 64-dimensional final layer features to classify CIFAR-100 classes during pretraining, but these classifier weights are ignored for supervised transfer. All other weights in the network are used to set the corresponding weights on the network to be trained for the target task. For SUN (397 classes x 5 images per class), we found it beneficial to finetune features by reducing the learning rate for the pretrained layers by a factor of 0.1 compared to the full learning rate used to train the 397-way classifier on top. For PASCAL-10 (10 classes x 5 images per class), only the 10-way action classifier was trained starting from random weights, while the weights of lower layers were frozen to their pretrained values, since finetuning was found to adversely impact classification results.

\vspace{-0.1in}
\paragraph{Dataset sample images}
Some sample images of KITTI, SUN, HMDB-51 and PASCAL-10 are shown at the end of this document.

\pagebreak

\clearpage
\begin{figure*}[h]
  \centering
  \includegraphics[width=1\linewidth]{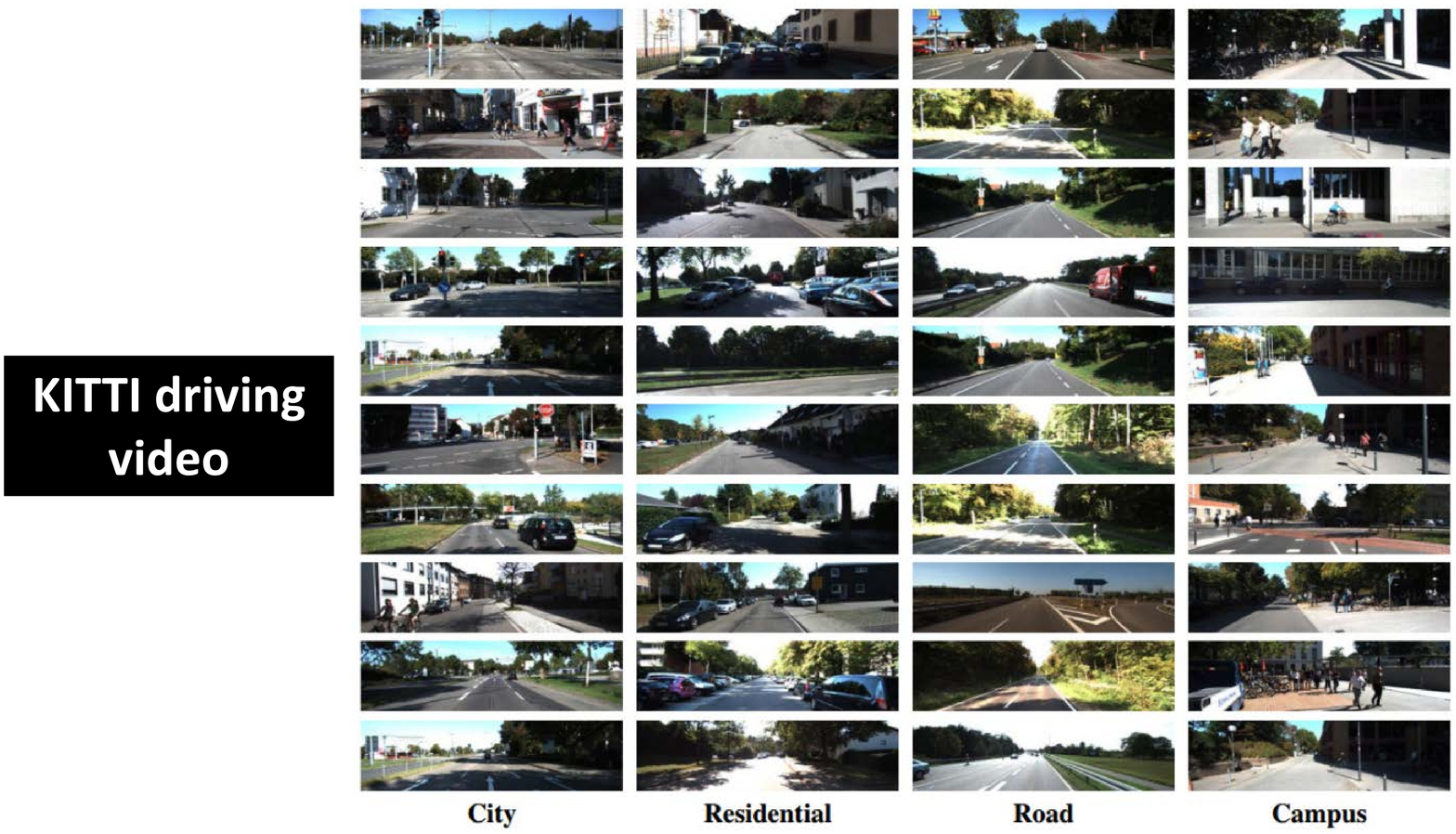}
  \label{fig:kitti}
\end{figure*}
\clearpage
\begin{figure*}[h]
  \centering
  \includegraphics[width=1\linewidth]{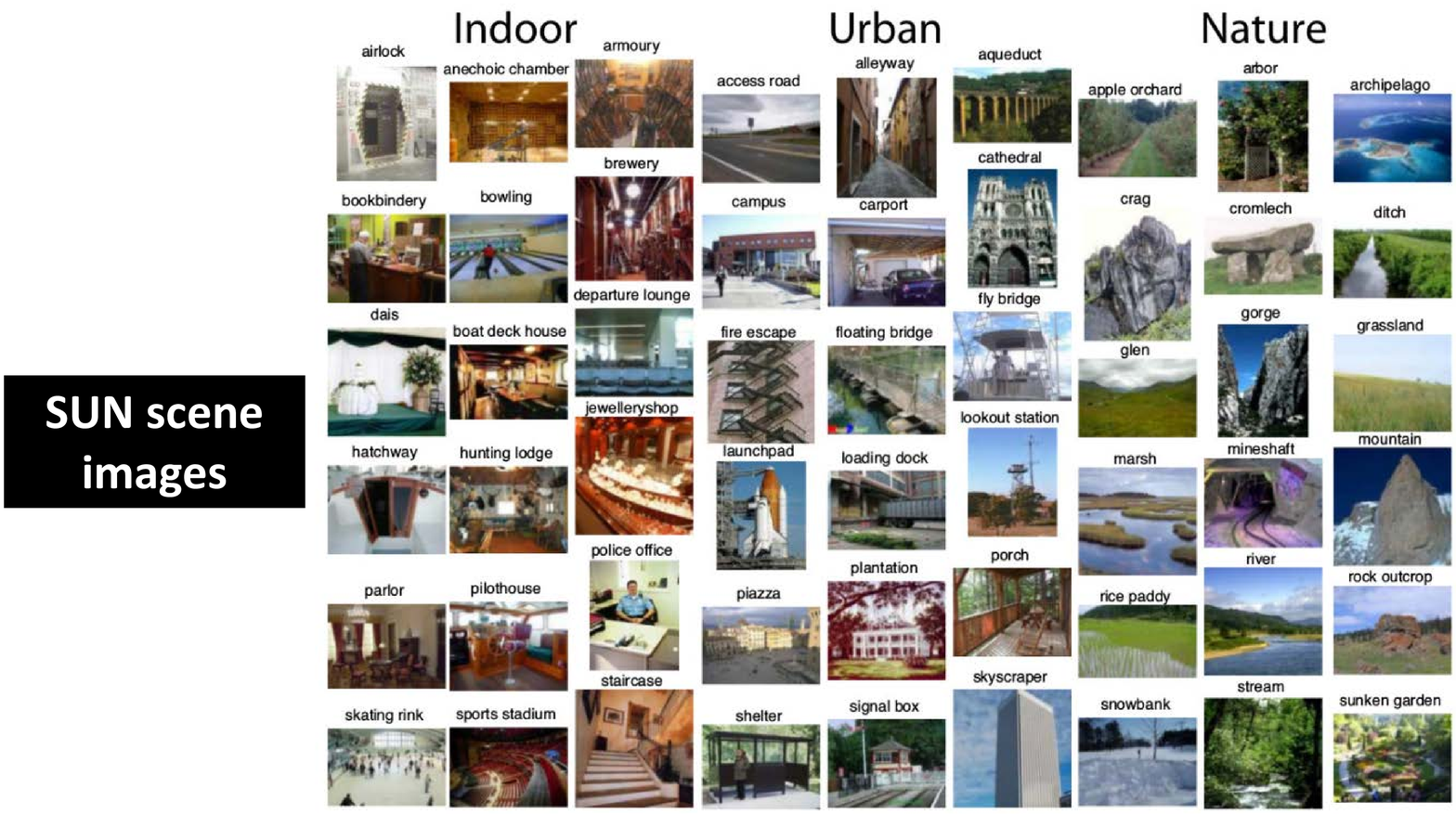}
  \label{fig:sun}
\end{figure*}
\clearpage
\begin{figure*}[h]
  \centering
  \includegraphics[width=1\linewidth]{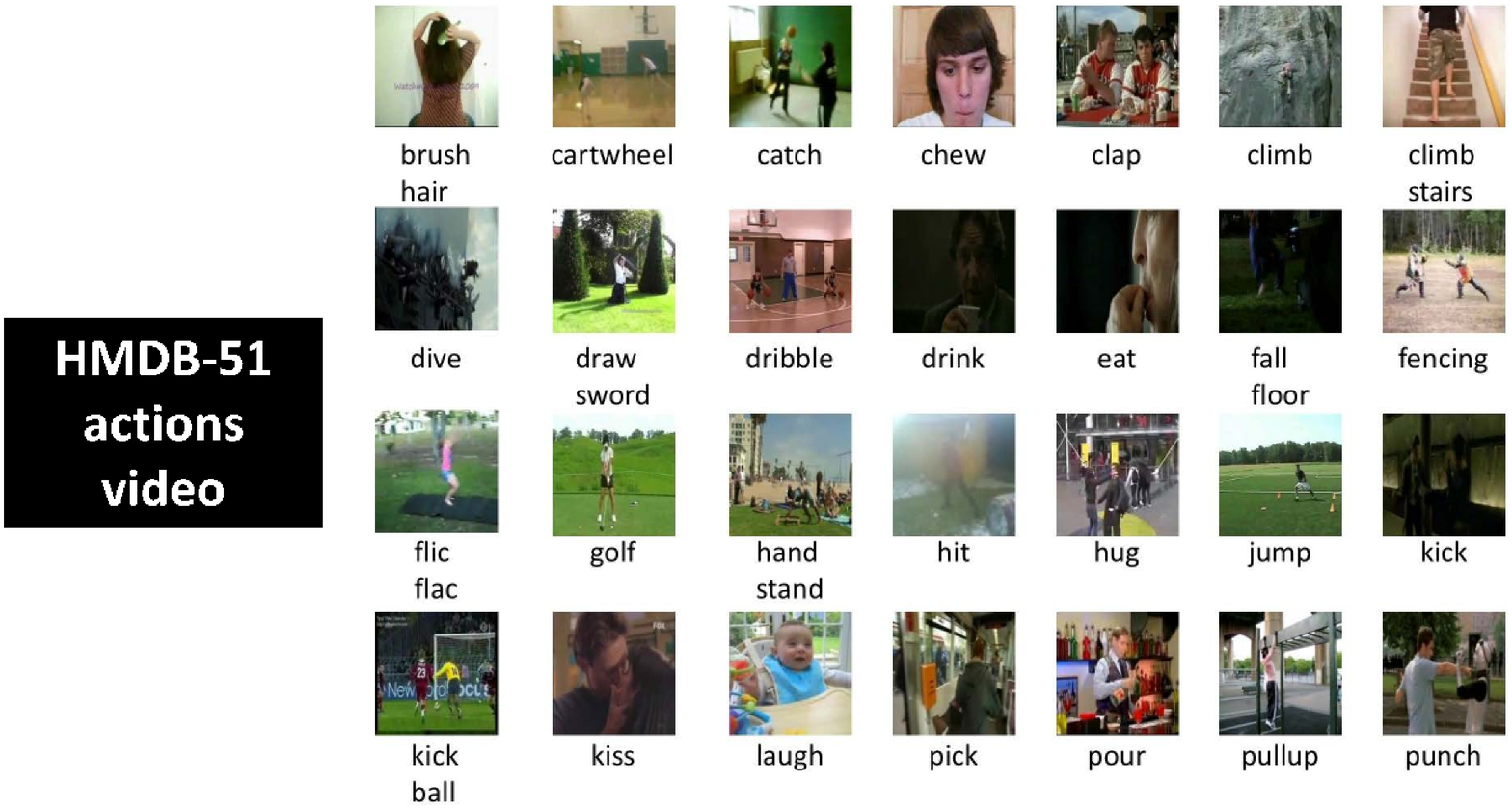}
  \label{fig:hmdb}
\end{figure*}
\clearpage
\begin{figure*}[h]
  \centering
  \includegraphics[width=1\linewidth]{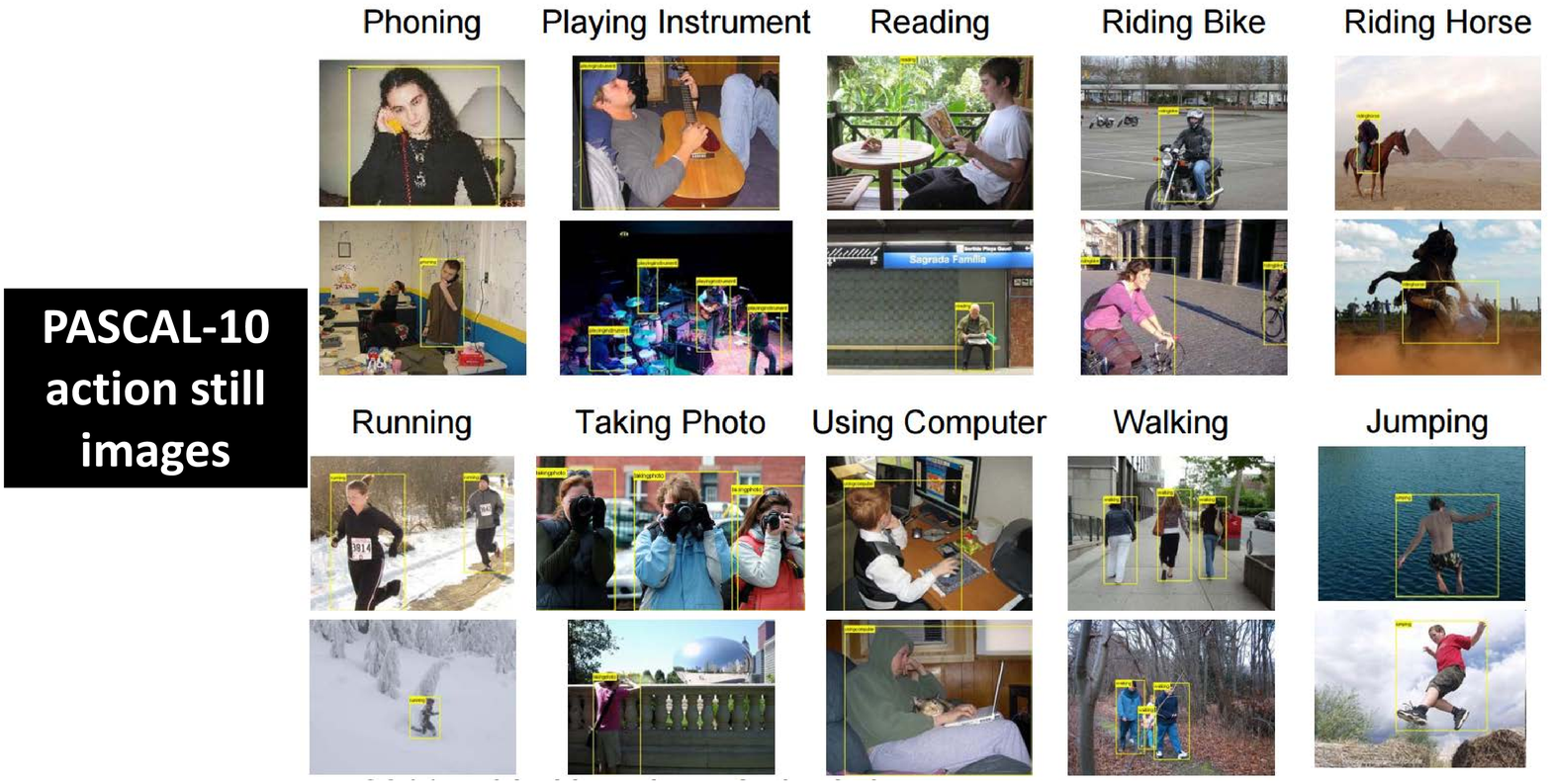}
  \label{fig:pascal}
\end{figure*}

\end{document}